%% file: miccai2020.tex
\documentclass[runningheads]{llncs}
\usepackage{graphicx}

\usepackage{algorithm}
\usepackage{algpseudocode}
\usepackage{hyperref}
\usepackage{xcolor}
\usepackage{amsmath}
\usepackage{color, colortbl}
\usepackage{diagbox}
\usepackage{subcaption}
\usepackage{cite}
\usepackage{multirow}
\definecolor{Gray}{gray}{0.95}
\algnewcommand{\LeftComment}[1]{\State \(\triangleright\) #1}

\newcommand\blfootnote[1]{%
  \begingroup
  \renewcommand\thefootnote{}\footnote{#1}%
  \addtocounter{footnote}{-1}%
  \endgroup
}

\begin{document}
\title{Inverse Distance Aggregation for Federated Learning with Non-IID Data} %
\author{Yousef Yeganeh\inst{1} \and Azade Farshad\inst{1} \and Nassir Navab\inst{1,2} \and Shadi Albarqouni\inst{1,3}
}
\authorrunning{Yeganeh et al.}
\institute{Computer Aided Medical Procedures, Technical University of Munich, Germany \and Whiting School of Engineering, Johns Hopkins University, United States \and Department of Computing, Imperial College London, United Kingdom}
\maketitle              %
\begin{abstract}
Federated learning (\textit{FL}) has been a promising approach in the field of medical imaging in recent years. A critical problem in \textit{FL}, specifically in medical scenarios is to have a more accurate shared model which is robust to noisy and out-of distribution clients. In this work, we tackle the problem of statistical heterogeneity in data for \textit{FL} which is highly plausible in medical data where for example the data comes from different sites with different scanner settings. We propose \textbf{IDA} (\textbf{I}nverse \textbf{D}istance \textbf{A}ggregation), a novel adaptive weighting approach for clients based on meta-information which handles unbalanced and non-iid data. We extensively analyze and evaluate our method against the well-known \textit{FL} approach, Federated Averaging as a baseline.

\blfootnote{Project page: \url{https://ida-fl.github.io/}}
\end{abstract}
\input{Intro.tex}
\input{Method.tex}
\input{Results.tex}
\input{Conclusion.tex}
\bibliographystyle{splncs04}
\bibliography{ref}

\end{document}

%% file: Intro.tex
\section{Introduction}
Federated learning (\textit{FL}) was proposed as a decentralized learning scheme where the data in each client is private and not exposed to other participants, yet they contribute to generation of a shared (global) model in a server that represents the clients' data~\cite{konevcny2015federated}. An aggregation strategy in the server is essential in \textit{FL} for combining the models of all clients. 
Federated Averaging (\textit{FedAvg})~\cite{mcmahan2016communication} is one of the most well-known \textit{FL} methods which uses the normalized number of samples in each client to aggregate the models in the server. Another aggregation approach using temporal weighting along with a synchronous learning strategy was proposed in ~\cite{chen2019communication}. Many recent approaches have been proposed in order to improve the generalization or personalization of the global model using the ideas of knowledge transfer, knowledge distillation, multi-task learning and meta-learning~\cite{li2019fedmd, jeong2018communication, smith2017federated, corinzia2019variational, chen2018federated, beel2018federated, jiang2019improving}.

Even though \textit{FL} has emerged into a promising and popular method to engage with privacy preserving distributed learning, it has faced some challenges: ~\textbf{a)} Expensive communication, ~\textbf{b)} privacy, ~\textbf{c)} systems heterogeneity and ~\textbf{d)} statistical heterogeneity\cite{li2019federated}. Although a large number of recent works on \textit{FL} such as ~\cite{sattler2019robust, liang2020think} are focused on communication efficiency due to its application on edge devices with unstable connections~\cite{li2019federated}, commonly using approaches such as compressed networks or compact features, its most determining aspects in the medical field are data privacy and heterogeneity~\cite{rieke2020future,kaissis2020secure}. Data heterogeneity assumption includes: ~\textbf{a)} Massively distributed: The data points are distributed among a very large number of clients. ~\textbf{b)} Non-iid (Not independent and identically distributed): Data in each node comes from a distinct distribution. The local data points are not representative of the whole data distribution (combination of all clients' data).~\textbf{c)} Unbalancedness: The number of samples across clients has a high variance. Such heterogeneity is foreseeable in medical data due to many reasons, for example, class imbalance in pathology, intra-/inter-scanner variability (domain shift), intra-/inter-observer variability (noisy annotations),  multi-modal data, and different tasks for clients.

There has been numerous works to handle each of these data assumptions~\cite{kairouz2019advances}. Training a global model with \textit{FL} in non-iid data is a challenging task. Model training in deep neural network suffers quality loss and may even diverge given non-iid data ~\cite{hsieh2019non}. There has been multiple works dealing with this problem. Sattler et al.~\cite{sattler2019clustered} propose clustering loss terms and using cosine similarity to overcome the divergence problem when clients have different data distributions.
Zhao et al. ~\cite{zhao2018federated} overcome the non-iid problem by creating a subset of data which is shared globally with the clients. In order to maintain system heterogeneity (affected by their main idea of nonuniform local updates), FedProx~\cite{li2018federated} proposes a proximal term to minimize the distance between the local and global models. Close to our approach, geometric median is used in~\cite{pillutla2019robust} to decrease the effect of corrupted gradients on the federated model.

In the last few years, there has been a growing interest in applying \textit{FL} in healthcare, in particular, to medical imaging.  Sheller et al.~\cite{sheller2018multi} were among the first works who applied \textit{FL} to multi-institutional data for Brain Tumor Segmentation task. 
To date, there has been numerous works on \textit{FL} in Healthcare~\cite{xu2019federated, li2019privacy, huang2019patient, li2020multi, sheller2020federated}. 
However, little attention has been paid to the aggregation mechanism given the data and system heterogeneity; for example, when the data is non-iid, or the participation rate of the clients is pretty low.

\begin{figure}[!hbtp]
\centering
\includegraphics[width=\textwidth]{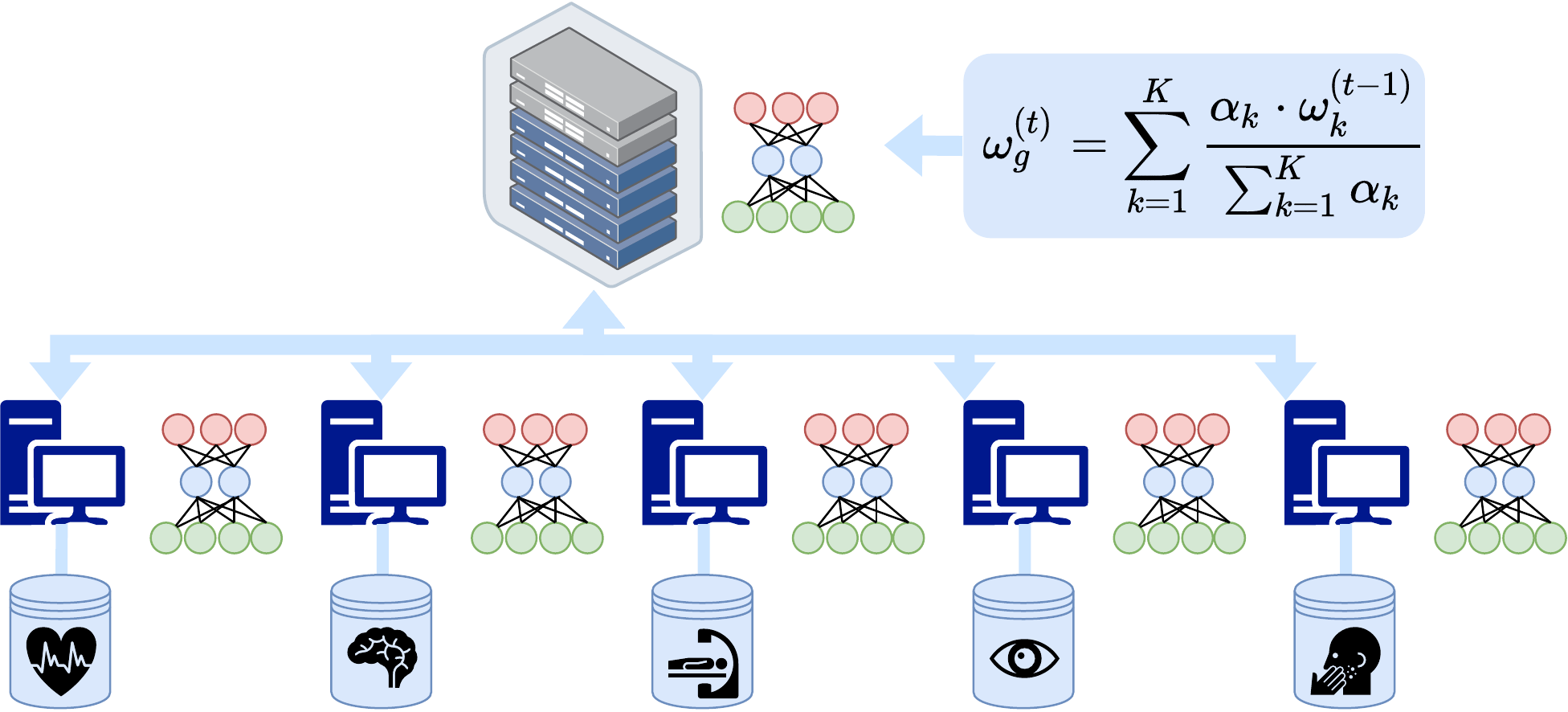}
\caption{Federated learning with non-iid data - The data has different distributions among clients.} \label{fig:teaser}
\end{figure}

In this work, we try to overcome the challenges of statistical heterogeneity in data and propose a robust aggregation method at the server side (\emph{cf.}~\autoref{fig:teaser}).
Our weighting coefficients are based on the meta-information extracted from the statistical properties of the model parameters. Our goal is to train a low variance global model given high variance local models which is robust to non-iid and unbalanced data. Our contributions are twofolds; ~\textbf{a)} A novel adaptive weighting scheme for federated learning which is compatible with other aggregation approaches, ~\textbf{b)} Extensive evaluation of different scenarios on non-iid data on multiple datasets.

Next, a brief overview of the federated learning concept is introduced in the methodology section before diving into the main contribution of the paper, the Inverse Distance Aggregation (IDA). Experiments and results on both machine learning datasets (Proof-of-Concept), and clinical use-cases are demonstrated and discussed.

%% file: Method.tex
\section{Method}
Given a set of $K$ clients with their own data distribution $p_k(x)$ and a shared neural network with parameters $\omega$, the objective is to train a global model minimizing the following objective function; 
\begin{equation}
    arg\min_{\omega_g^t} f(x; \omega_g^t), \hspace{10pt} \text{where} \hspace{10pt} f(x; \omega_g^t) = \sum_{k=1}^K f(x; \omega_k^t), 
\end{equation}
where $\omega_g^t, \omega_k^t$ are the global and local parameters, respectively. 

\subsection{Client}

Each randomly sampled client, from the total number of $K$ clients (based on the participation rate $pr$), receives the global model parameter $\omega_g^t$ at communication round $t$, and trains the shared model, initialized by $\omega_g^t$, on its own training data $p_k(x)$ for $E$ iterations to minimize its local objective function $f_k (x) = \mathop{\mathbf{E}}_{x \sim p_k(x)}[f(x; \omega_k^t)]$ where $\omega_k^t$ is the weight parameters of the client $k$. The training data in each client is a subset of the whole training data, which can be sampled from different classes of data. The number of classes of data assigned to each client is denoted by $n_{cc}$.

\subsection{Server}
Each round $t$, the updated local parameters $\omega_k^{t}$ are sent back to the server and aggregated to form the updated global parameter $\omega_g^{t}$,  
\begin{equation}
    \omega_g^t = \sum_{k=1}^K \alpha_k \cdot \omega_k^{t-1}.
\label{eq:aggr}
\end{equation} 
where $\alpha_k$ is the weighting coefficient. This procedure continues for the given total communication rounds $T$.

\subsection{Inverse Distance Aggregation (IDA)}
In order to reduce the inconsistency among the updated local parameters due to the non-iid problem, we propose a novel robust aggregation method, denoted as Inverse Distance Aggregation (\textbf{IDA}). 
The core of our method is the way the coefficients $\alpha_k$ are computed, which is based on the inverse distance of each client parameters to the average model of all clients. This allows us to reject or weigh less the models who are poisoning, \emph{i.e.} out-of-distribution models.

To realize this, the $\ell_1$-norm is utilized as a metric to measure the distance of clients $\omega_{k}$ to the average one $\omega_{Avg}$ as 
\begin{equation}
\alpha_{k} = 
\frac{1}{Z}\| \omega^{t-1}_{Avg} - \omega^{t-1}_{k} \|^{-1},
\label{eq:IDA}
\end{equation}
where $Z = \sum_{k \in K} \| \omega^{t-1}_{Avg} - \omega^{t-1}_{k} \|^{-1}$ is a normalization factor. In practise, we add $\epsilon$ to both numerator and denominator to avoid any numerical instability. Note that $\alpha{_k} = 1$ when clients' parameters is equivalent to the average one, and $\alpha{_k} = n_k$ is equivalent to the \textit{FedAvg}~\cite{mcmahan2016communication}.

We also propose to use the training accuracy of clients in the final weighting which we denote by INTRAC (INverse TRaining ACcuracy) to penalize over-fitted models and encourage under-trained models in the aggregated model. To calculate the coefficients for INTRAC, We assign $\alpha^\prime{_k} = \frac{Z^\prime}{max(\frac{1}{K}, acc_k)}$. The $max$ function is used to assure all of the values are above chance level. Here $acc_k$ is the training accuracy of client $k$, $\alpha^\prime{_k}$ is the INTRAC coefficient and $Z^\prime = \sum_{k \in K} max(\frac{1}{K}, acc_k)$ is the normalization factor. We normalize the calculated coefficients $\alpha^\prime{_k}$ once again to bring them to the range of $(0,1]$. To combine different coefficient values (i.e. INTRAC, IDA, FedAvg), we multiply the acquired coeffecients and normalize them in the range of $(0,1]$.

%% file: Results.tex
\section{Experiments and Results}
We evaluated our method on commonly used databases to show a Proof-of-Concept (PoC) before we present some results on a clinical use-case. We compare the results of our method \textbf{IDA} against the baseline method \textit{FedAvg}~\cite{mcmahan2016communication}. In the first set of PoC experiments, we investigate the following: 1) Non-iid vs. iid: Comparison of \textit{FedAvg} and \textbf{IDA} in iid and non-iid with different datasets and architectures.
2) Ablation study: Investigation of effectiveness of IDA compared to FedAvg
3) Sensitivity analysis: Performance comparison in extreme situations.

\paragraph{Datasets} We show the results of our evaluation on cifar-10~\cite{krizhevsky2009learning}, fashion-mnist (f-mnist)~\cite{xiao2017fashion} and HAM10K(multi-source dermatoscopic images of pigmented lesions)\cite{tschandl2018ham10000} datasets. f-mnist is a well-known variation of mnist with $50k$ images of $28\times 28$ black and white clothing pieces. cifar-10 is another dataset with $60k$ $32\times 32$ images of vehicles and animals, commonly used in computer vision. For the clinical study, we evaluate our method on HAM10k dataset which includes a total number of $10015$ images of different pigmented skin lesions in $7$ classes. The different classes and their number of samples in HAM10k are as follows: Melanocytic nevi: 6705, Melanoma: 1113, Benign keratosis: 1099, Basal cell carcinoma: 514, Actinic Keratoses: 327, Vascular: 142, Dermatofibroma: 115. We chose this dataset due to its heavy unbalancedness.

\paragraph{Implementation Details}
The training settings for each dataset are: LeNet\cite{lecun1989backpropagation} for f-mnist with 10 classes, batchsize=128, learning rate (lr)=0.05 and local iteration of 1 (E=1), VGG11~\cite{simonyan2014very} without batch normalization and dropout layers for cifar-10 with 10 classes and batchsize=128, lr=0.05 and E=1. For HAM10K, we used Densenet-121~\cite{huang2017densely} with 7 classes, batchsize=32, lr=0.016 and E=1. In all of the experiments $90\%$ ofr the images are randomly sampled for training and the rest are employed for evaluation. All of the models are trained for a total number of $5000$ rounds. The mentioned values are the default for all experiments unless otherwise specified.

\paragraph{Evaluation Metrics}
In all of the experiments, we separate a part of each client's dataset as its test set, and we report the accuracy of the global (aggregated) model on the union of the test sets of clients and the local accuracy of each client on it's own local test data. This gives us an indication of how well the global model is representative of the aggregated dataset. We report the classification accuracy in all of the experiments.

\subsection{Proof-of-Concept}
\subsubsection{Non-iid vs. iid}
In this section we evaluate and compare \textbf{IDA} with \textit{FedAvg} on f-mnist and cifar-10 datasets given different scenarios of data distribution in clients. ~\autoref{tab:iidness} demonstrates the results of balanced data distribution where all clients have the same or similar number of samples for $n_{cc} \in \{3,5, 10 (iid)\}$ and $pr \in \{30\%,50\%,100\%\}$. Our results show that \textbf{IDA} has slightly better  or on-par performance to \textit{FedAvg} in all scenarios of balanced data distribution. 
\newcolumntype{g}{>{\columncolor{Gray}}c}
\begin{table}[t]
\caption{Comparison between our method and the baseline on cifar10 and f-mnist with different number of classes per client in non-iid and iid scenarios}\label{tab:iidness}
\resizebox{\textwidth}{!}{
\footnotesize
\centering
\begin{tabular}{|g|g|l|l|l|l|l|l|l|}
\hline
\rowcolor{Gray}

\hline
&$n_{cc}$     & \multicolumn{3}{|c|}{3c}      & \multicolumn{3}{c|}{5c}                          & \multicolumn{1}{c|}{iid} \\ \cline{2-9}
\rowcolor{Gray}
\multirow{-2}{*}{Dataset} &\diagbox[width=2.4cm,height=0.6cm]{Method}{$pr$} & 30\% & 50\% & 100\% & 30\%        & 50\%        & 100\%       & 100\%                 \\ \hline
&FedAvg &  63.20      & 65.11   &  69.81    & 19.68          & 83.11          & 80.94          & 87.77                    \\ \cline{2-9}
\multirow{-2}{*}{cifar-10}&IDA    & \textbf{64.36}        & \textbf{67.70}   & \textbf{70.80}    & \textbf{76.06} & \textbf{83.55} & \textbf{83.82} & \textbf{89.46}           \\ \hline
&FedAvg & 86.23          & 87.09          & \textbf{87.45} & 87.60          & 87.81          & 87.16          & 86.95                    \\ \cline{2-9}
\multirow{-2}{*}{f-mnist}&IDA    & \textbf{87.64} & \textbf{87.61} & 87.44          & \textbf{87.93} & \textbf{87.89} & \textbf{87.46} & \textbf{87.10}            \\ \hline
\end{tabular}
}
\end{table}

\subsubsection{Ablation Study}
In this section, we investigate the effect of different components of the weighting coefficients. We evaluate all of the proposed components on cifar-10 and f-mnist and compare them with two baseline methods, namely \textit{FedAvg}, and another baseline where $\alpha_k =1$, denoted by \textit{Mean} shown in ~\autoref{tabs:ablationv1}. We also evaluate the combination of our weighting method with number of samples per client (\textbf{IDA + FedAvg}) and adding the training accuracy of each client to the weighting scheme (\textbf{IDA + INTRAC}). The results indicate that combining different weighting schemes can lead to a better performing global model in FL. This supports our hypothesis, that if some of the clients have lower quality or poisonous models, \textit{FedAvg} would be vulnerable, but our methods can lower the contribution of bad models (overfitted, low quality or poisonous models) so the final model performs better on the federated dataset.

\begin{table}[t]
\caption{Ablation study on different weighting combinations on f-mnist and cifar-10 datasets.}\label{tabs:ablationv1}
\resizebox{\textwidth}{!}{
\footnotesize
\centering
\begin{tabular}{|g|>{\centering\arraybackslash}p{4cm}|>{\centering\arraybackslash}p{4cm}|>{\centering\arraybackslash}p{4cm}|}
\hline
\rowcolor{Gray}
\hline
       & f-mnist $|$ $n_{cc}=3$ $|$ pr=30\% & \multicolumn{2}{|c|}{cifar-10 $|$ $n_{cc}=3$ $|$ $pr=30\%$}  \\ \hline
\rowcolor{Gray}
\multirow{-2}{*}{\diagbox[width=2.4cm,height=0.6cm]{Method}{Settings}}      & K=10 & K=10 & K = 20 \\ \hline
Mean       &  87.47                   & 65.82       & 84.80 \\ \hline
FedAvg     & 86.23                   & 63.20          & 22.84 \\ \hline
IDA        & 87.64                   & 64.36            & 83.98 \\ \hline
IDA + FedAvg & 86.67                   & \textbf{67.29}          & 82.14 \\ \hline
IDA +INTRAC & \textbf{88.33}          & 64.93   & \textbf{85.23} \\ \hline
\end{tabular}
}
\end{table}

\subsubsection{Sensitivity analysis}
In real-life scenarios, stability of learning process in unfavorable conditions is critical. In \textit{FL} it is not mandatory for the members to contribute in each round, so the participation rate can be different in each round of training, and we might have lower quality models in any round. It is very likely that some clients have very few data samples, and some other clients have a lot of data. In this section we investigate the global model's performance given low participation rate and severe non-iidness.

\begin{figure*}[h]
    \centering
    \begin{subfigure}[t]{0.49\textwidth}
        \centering
        \includegraphics[width=\textwidth]{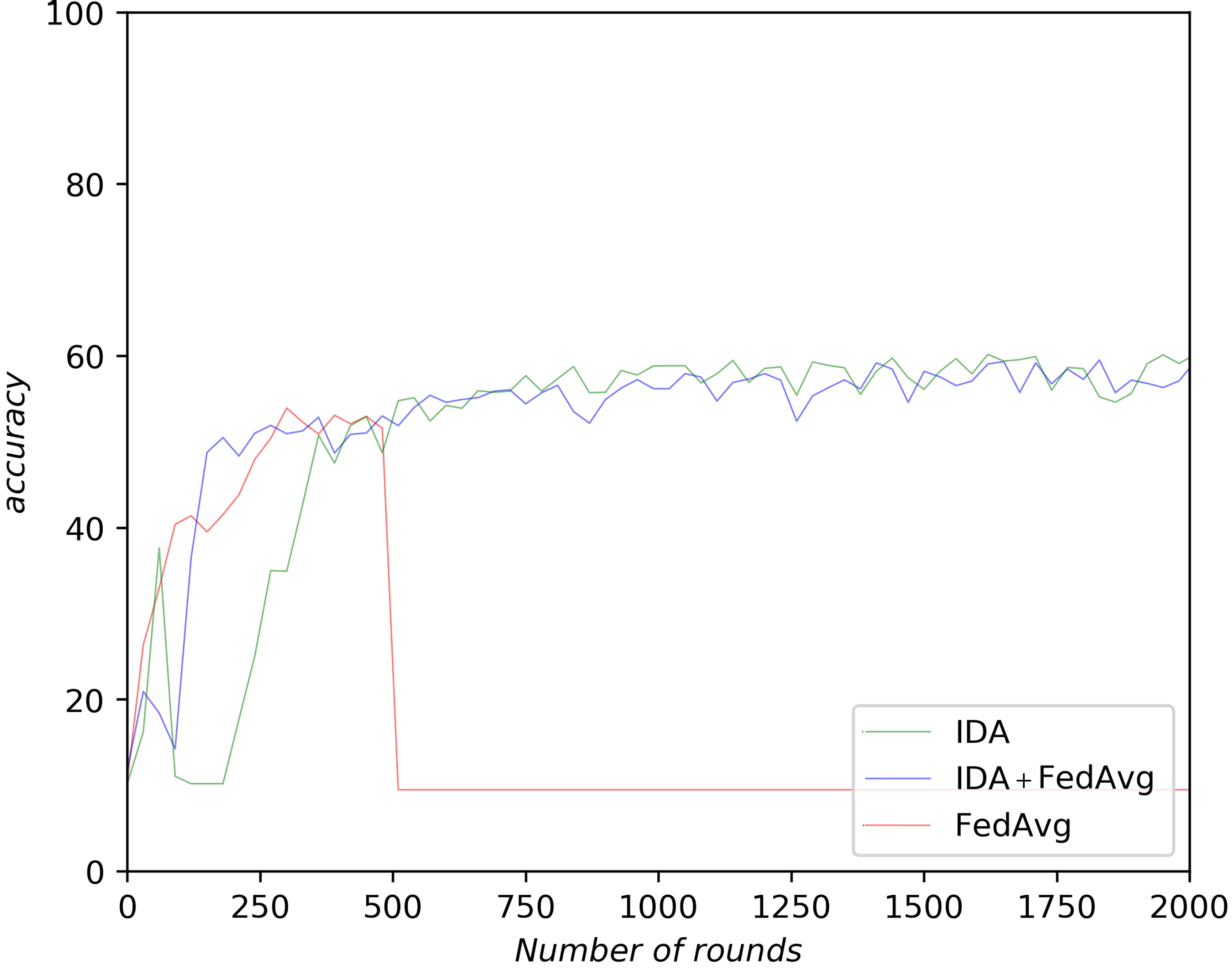}
        \label{figs:pr_effect_l}
    \end{subfigure}%
    \begin{subfigure}[t]{0.49\textwidth}
    \centering
    \includegraphics[width=\textwidth]{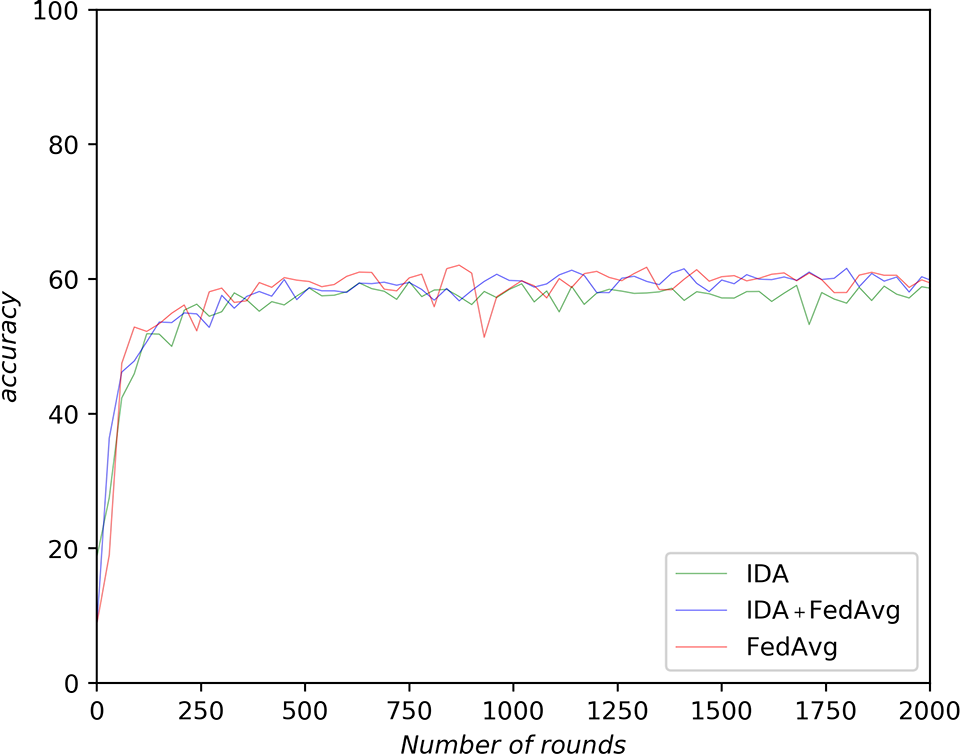}
    \label{figs:pr_effect_r}
    \end{subfigure}
    \caption{Left: participation rate (pr) of 0.01; Right: participation rate of 0.05. The pr affects the stability of federated learning, and it is shown that \textbf{IDA} has stable performance comparing to FedAvg.}
    \label{figs:lr_comparison}
\end{figure*}

\paragraph{Low participation rate in non-iid distribution}
To investigate the effect of participation rate, we used 1000 clients on f-mnist dataset with (batchsize=30, $lr=0.016$ and $n_{cc}=3$ and each client has up to $500$ samples). In this experiment, we observe that despite the fact that this dataset is relatively easy to learn, decreasing the participation rate of clients lowers the performance (\emph{cf.}~\autoref{figs:lr_comparison}). When the participation rate is at $1\%$, the model trained using \textit{FedAvg} collapses. However, when we increase the participation rate to $5\%$ the model continues to learn. We observe a robust performance for both \textbf{IDA} and \textbf{IDA + FedAvg} in both scenarios.

\begin{figure*}[t]
    \centering
    \begin{subfigure}[t]{0.49\textwidth}
        \centering
        \includegraphics[width=\textwidth]{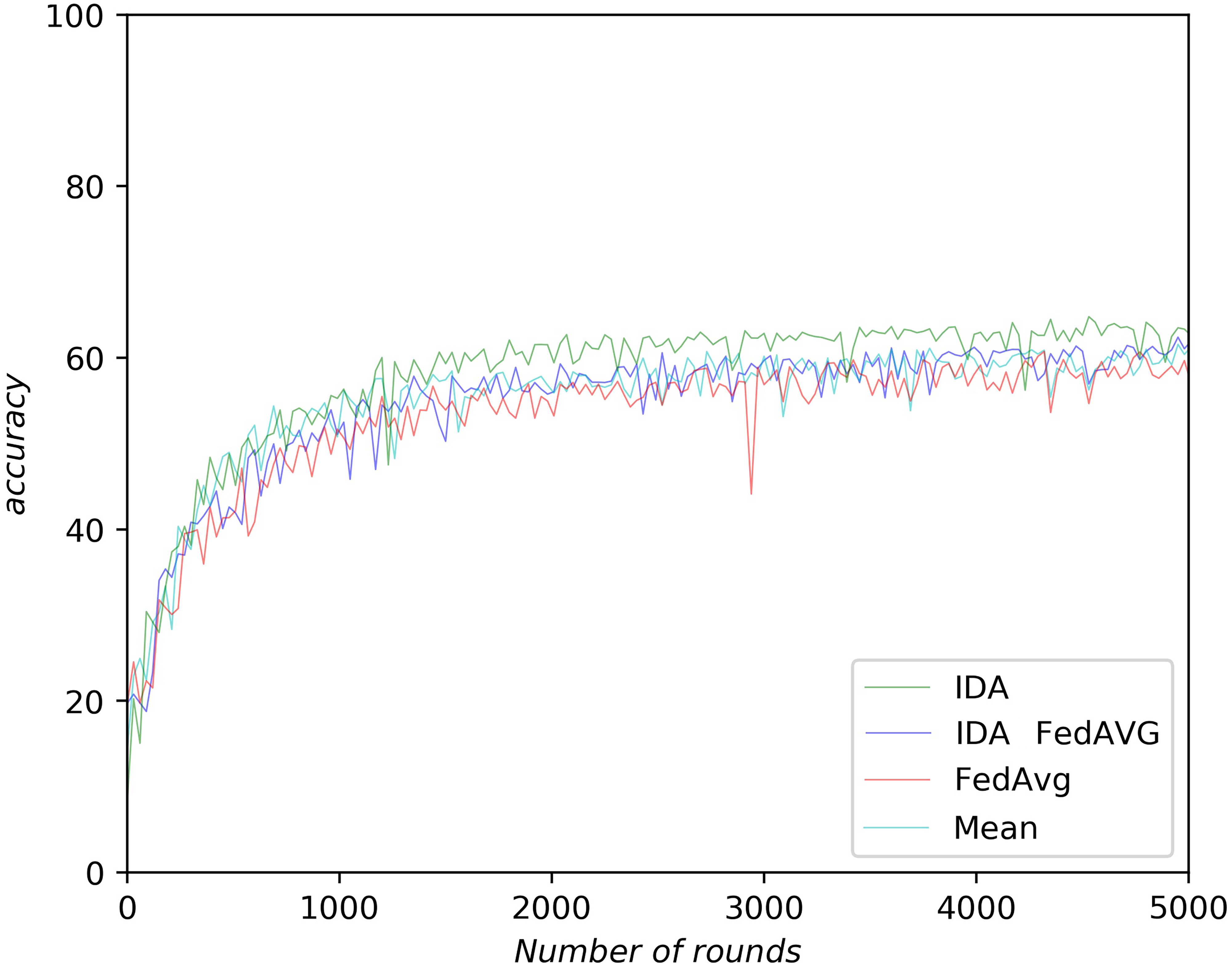}
        \label{figs:unbl_proof_l}
    \end{subfigure}%
    \begin{subfigure}[t]{0.49\textwidth}
        \centering
        \includegraphics[width=\textwidth]{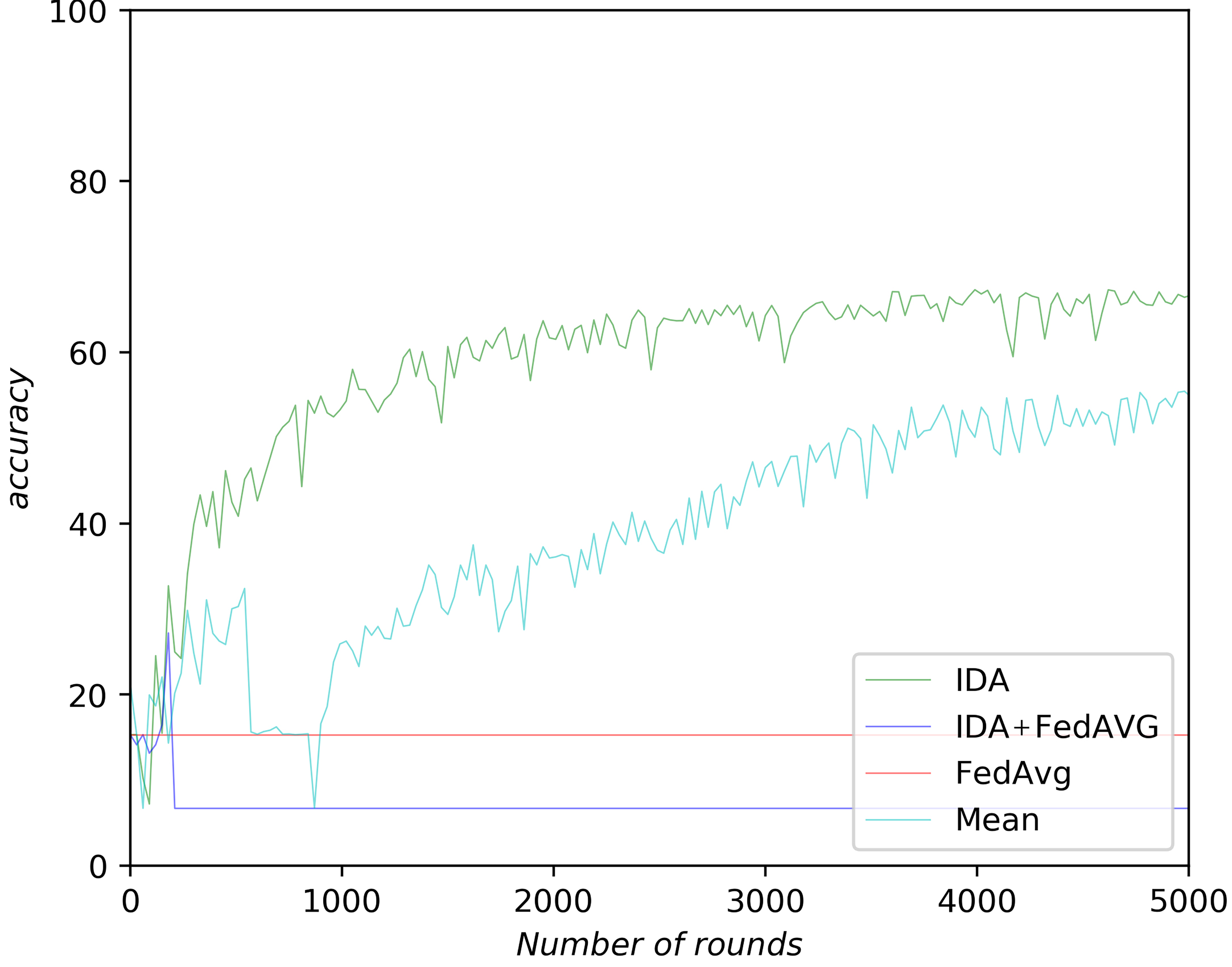}
        \label{figs:unbl_proof_r}
    \end{subfigure}
    \caption{Accuracy of global model of clients with non-iid data distribution on cifar-10: in the right we have the same clients, and the same learning hyperparameters of the left, but the number of samples in three of the clients with poor performances increased. The local distribution of data points in those three clients remained the same. This experiment is performed on cifar-10 dataset with $K=10$ clients, $n_{cc}=3$, $E=2$, lr=0.01 and  random number of samples per class per client up to 1000 samples.}
    \label{figs:compare_noniid}
\end{figure*}

\paragraph{Severity of Non-IID}
To analyze the effect of non-iidness on the performance of our method, we design an experiment by increasing the data samples of the low performing clients.
To achieve this, first we train our models in a normal fashion as mentioned in previous sections. Then we choose three clients with the lowest accuracy at the end of the initial training and double the amount of their samples in the training data distribution. We repeat the training using the newly generated data distribution. We propose this experiment to see the effect of \textit{FedAvg} weighting in a scenario where low performing clients are given higher weight. It can be seen in \autoref{figs:compare_noniid} that before increasing the number of samples, \textbf{IDA} performs marginally better compared to other methods; however, after we increase the number of samples in those three clients, \textit{FedAvg} collapses at the beginning of training. Considering the performance of \textit{Mean} aggregation, we see that \textbf{IDA} is the main contributing factor to the learning process.

\subsection{Clinical use-case}
We evaluate our proposed method on HAM10k dataset and show our results in ~\autoref{tab:ham10k}. Even though the global accuracy of the model using \textbf{IDA} is on par with \textit{FedAvg}, it can be seen that the local accuracy (accuracy of clients on their own test set) using \textbf{IDA} is superior to \textit{FedAvg} in all scenarios. This indicates that \textbf{IDA} has a better generalization and lower variance in local accuracy of clients.

\begin{table}[!h]
\caption{Investigation on an unbalanced data distribution among the clients in federated setting, with five random classes per client, and random number of samples per client for HAM10k.}\label{tab:ham10k}
\footnotesize
\centering
\begin{tabular}{|g|g|c|c|}
\rowcolor{Gray}

\hline
Method & $n_{cc}$            & Global Accuracy  & Local Accuracy                      \\ \hline
FedAvg                                 & $1$                 & $\textbf{69.72}$ & $       {60.52} \pm        {9.20}$   \\ \hline
IDA                                    & $1$                 & $69.16$          & $\textbf{61.21} \pm \textbf{8.79}$  \\ \hline
FedAvg                                 & $2$                 & $\textbf{62.23}$ & $57.14 \pm 10.84$                   \\ \hline
IDA                                    & $2$                 & $61.21$          & $\textbf{60.21} \pm \textbf{5.48}$  \\ \hline
FedAvg                                 & $10 \mbox{ (iid)} $ & ${63.5}$         & $52.88 \pm 15.73$  \\\hline  
IDA                                    & $10 \mbox{ (iid)} $ & $\textbf{63.72}$ & $\textbf{57.38} \pm \textbf{10.56}$ \\ \hline
\end{tabular}
\end{table}

%% file: Conclusion.tex
\section{Discussion and Conclusion}
In this work, we proposed a novel weighting scheme for aggregation of client models in a federated learning setting for non-iid and unbalanced data distribution. 
Our weighting is calculated based on the statistical meta-information which gives higher weights in aggregation to the clients that their data has a lower distance to the global average. We also propose another weighting approach called INTRAC that normalizes models to lower the contribution of overfitted models to the shared model. Our extensive experiments show that our proposed method outperforms FedAvg in terms of classification accuracy in non-iid scenario. Our proposed method is also resilient to low quality or poisonous data in the clients. For instance, if the majority of clients are rather aligned, then they can rule out the out-of-distribution models. This is not the case with FedAvg, however, which is based on the presumption that the clients with more data, have a better distribution compared to other models, and they should have more voting power in the global model. Future research directions concerning the out-of-distribution models detection and robust aggregation schemes should be further considered.  

\section*{Acknowledgements}
S.A. is supported by the PRIME programme of the German Academic Exchange Service (DAAD) with funds from the German Federal Ministry of Education and Research (BMBF). A.F. is supported by Munich Center for Machine Learning (MCML) with funding from the German Federal Ministry of Education and Research (BMBF) under Grant No. 01IS18036B. We also gratefully acknowledge the support of NVIDIA Corporation with the donation of the Titan V GPU used for this research.